\def\year{2019}\relax
\documentclass[letterpaper]{article} 
\pdfoutput=1
\usepackage{aaai18}  
\usepackage{times}  
\usepackage{helvet}  
\usepackage{courier}  
\usepackage{url}  
\usepackage{graphicx}  
 
\usepackage{amsmath}
\usepackage{amsfonts}
\usepackage{bm}
\usepackage{xcolor}
\usepackage{subcaption}
\usepackage{hyperref}

\begin{document}
%
\title{Causal Simulations for Uplift Modeling}
\author{Jeroen Berrevoets \and Wouter Verbeke\\
Vrije Universiteit Brussel\\
Data Analytics Laboratory\\
{\texttt{\{jeroen.berrevoets, wouter.verbeke\}@vub.be}}
}
\maketitle
\begin{abstract}
	Uplift modeling requires experimental data, preferably collected 
	in random fashion. This places a logistical and financial burden 
	upon any organisation aspiring such models. Once deployed, uplift 
	models are subject to effects from concept drift. Hence, methods 
	are being developed that are able to learn from newly gained 
	experience, as well as handle drifting environments. As these new 
	methods attempt to eliminate the need for experimental data, another 
	approach to test such methods must be formulated. Therefore, 
	we propose a method to simulate environments that offer causal 
	relationships in their parameters.
\end{abstract}

\section{Introduction} 
\noindent In uplift modeling, the effect $ E $ of a cause $ C $ applied 
on an entity $ \mathbf{x} $ is estimated. Effect is defined as a measure 
of behaviour. When this behaviour is paired with a higher probability of 
occurring after the application of $ C $, an uplift is associated. In essence, 
the uplift is thus the positive net impact of $ C $ on $ E $ for a specific 
$ \mathbf{x} $, as defined in (\ref{eq:uplift}). 
\begin{equation} \label{eq:uplift}
	\fontsize{9.0pt}{10.0pt} U(C, E, \mathbf{x}) 
	\doteq p(E = 1| C = 1, \mathbf{x}) - p(E = 1| C = 0, \mathbf{x})
\end{equation}
While above formulation describes the objective, it is impossible to 
derive from data due to the fundamental problem of causal inference 
\cite{holland1986}. As such, an estimation of the uplift $ \hat{u}(C, 
E, \bm{\phi}(\mathbf{x})) $ over a generalisation $ \bm{\phi}(\cdot) $ 
of the entities is needed. Typical applications of such models are
found in direct marketing \cite{devriendt2018}, as well as medical 
scenarios \cite{rzepakowski2012}. 

While the collection of experimental data in the context of direct marketing 
seems acceptable, the inability to handle concept drift is not \cite{fang2018,tsymbal2004}. 
Contrasting a medical setting where concept drift is likely of lesser concern.  
However, the randomised collection of experimental data could prove problematic. 
This due to the need for a sufficient amount of controlled -- or untreated -- cases. 

Alas, every model is learned on the basis of data. However, the method of 
gathering this data tends to quickly diverge from a purely random manner, 
given an experienced based learner as defined in the field of reinforcement 
learning (RL). 

While in RL some state could potentially transition to a subsequent state, 
in this first formalisation this consideration is omitted in favour of bandit 
algorithms \cite{robbins1952}. Research on causally aware bandits is reliant 
on either a real environment \cite{sawant2018,li2010}, or a simulation 
\cite{lee2018,sen2016,lattimore2016,bareinboim2015}. Depending on the 
projected use of the model, testing experimental algorithms on a real 
environment might prove detrimental. 

It is therefore crucial to formulate a variety of simulated environments to 
both validate and benchmark these algorithms. As such, a simulation offers the 
ability to consult a ground truth -- or counterfactual -- to properly measure 
performance. This research contributes a method by which such causal simulations 
are to be created. The method presented is highly generic and covers a wide range 
of situations a causal bandit could be faced with. 

\section{Simulating an environment}
When constrained with $ C \in \{0,1\} $, an uplift model provides an estimated 
difference between two conditional probabilities. If such a difference is absent,
$ C $ does not offer any causal relationships in the environment. Therefore, 
this difference is assumed. As such, any causal simulation should provide at 
least two distinct distributions; $ \mathcal{D}_0 $ for the controlled cases, 
and $ \mathcal{D}_1 $ for the treated cases where $ E \sim \mathcal{D}_i = p(E=1 
| C=i, \mathbf{x}) \in \mathbb{D} $ as in (\ref{eq:uplift}). 

When $ k $ causes are to be modeled, the simulation ought to provide $ k+1 $ 
distributions. A difference could be caused by the interaction of unobserved 
confounders \cite{bareinboim2015}. Any unobserved confounder will introduce a 
dimension to $ \mathcal{D}_i $ and is discussed in a following section.

\begin{figure*}[h!]
	\centering
	\begin{subfigure}[t]{0.3\textwidth}		
        \centering
        \includegraphics[height=1.2in]{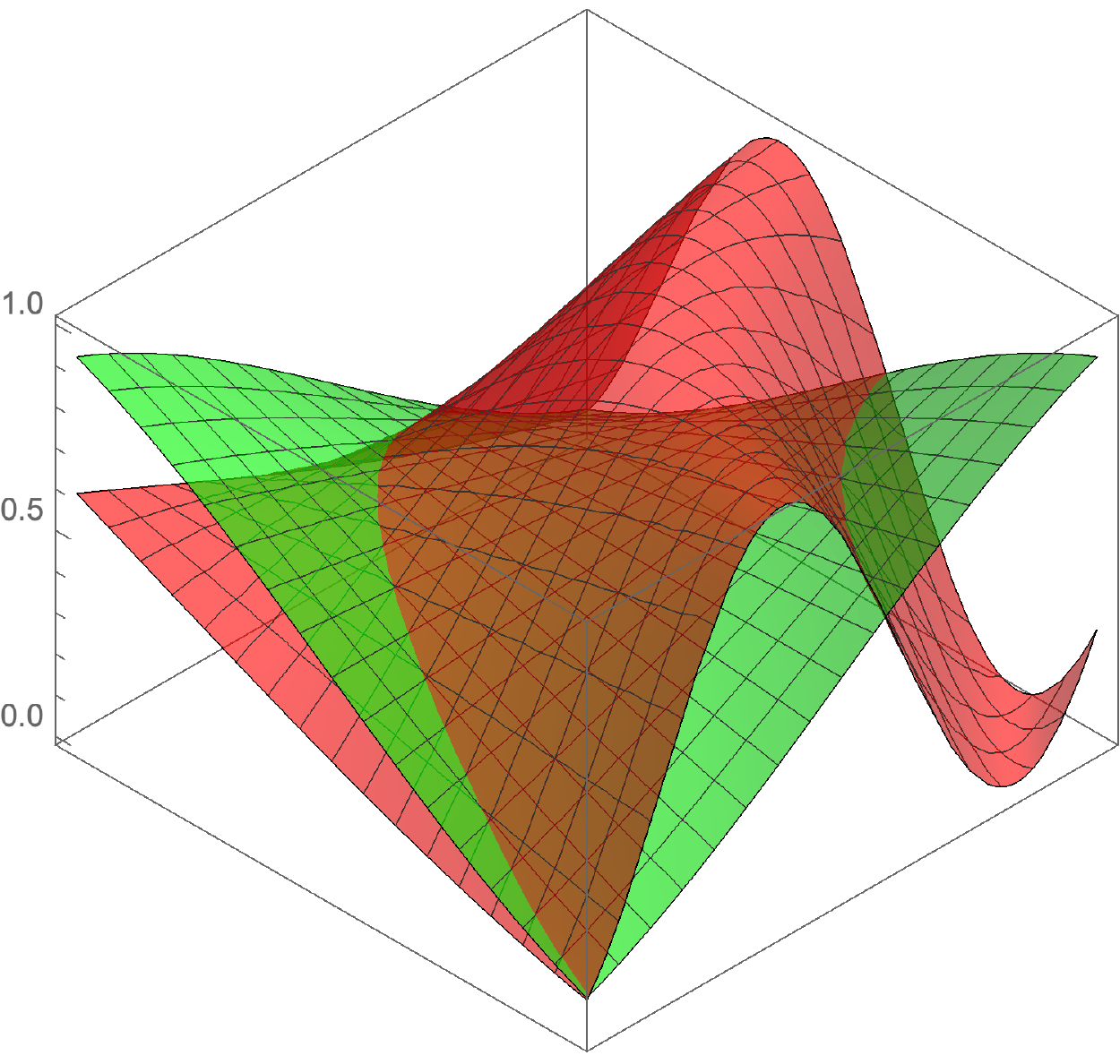}
		\caption{Sine base, with $ \lambda(C=0)=2 $, $ \lambda(C=1)=1 $ 
		and $ \mathbf{g} = (0.7, 0.7)^T $}
		\label{fig:sine}
    \end{subfigure}%
    ~ 
	\begin{subfigure}[t]{0.3\textwidth}		
        \centering
        \includegraphics[height=1.2in]{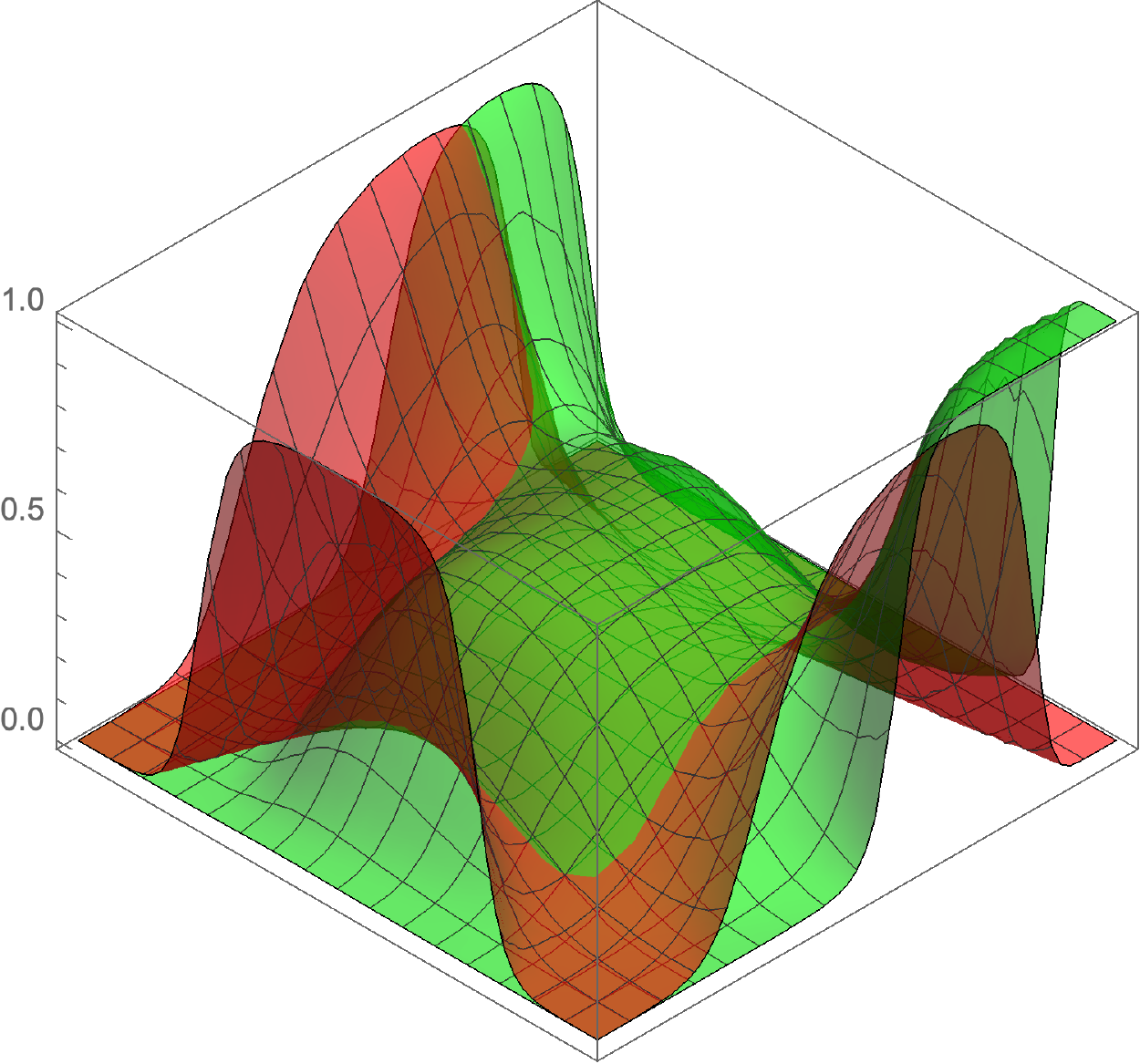}
		\caption{Polynomial base, with $ q=5 $}
		\label{fig:poly}
	\end{subfigure}
	~
	\begin{subfigure}[t]{0.3\textwidth}		
        \centering
        \includegraphics[height=1.2in]{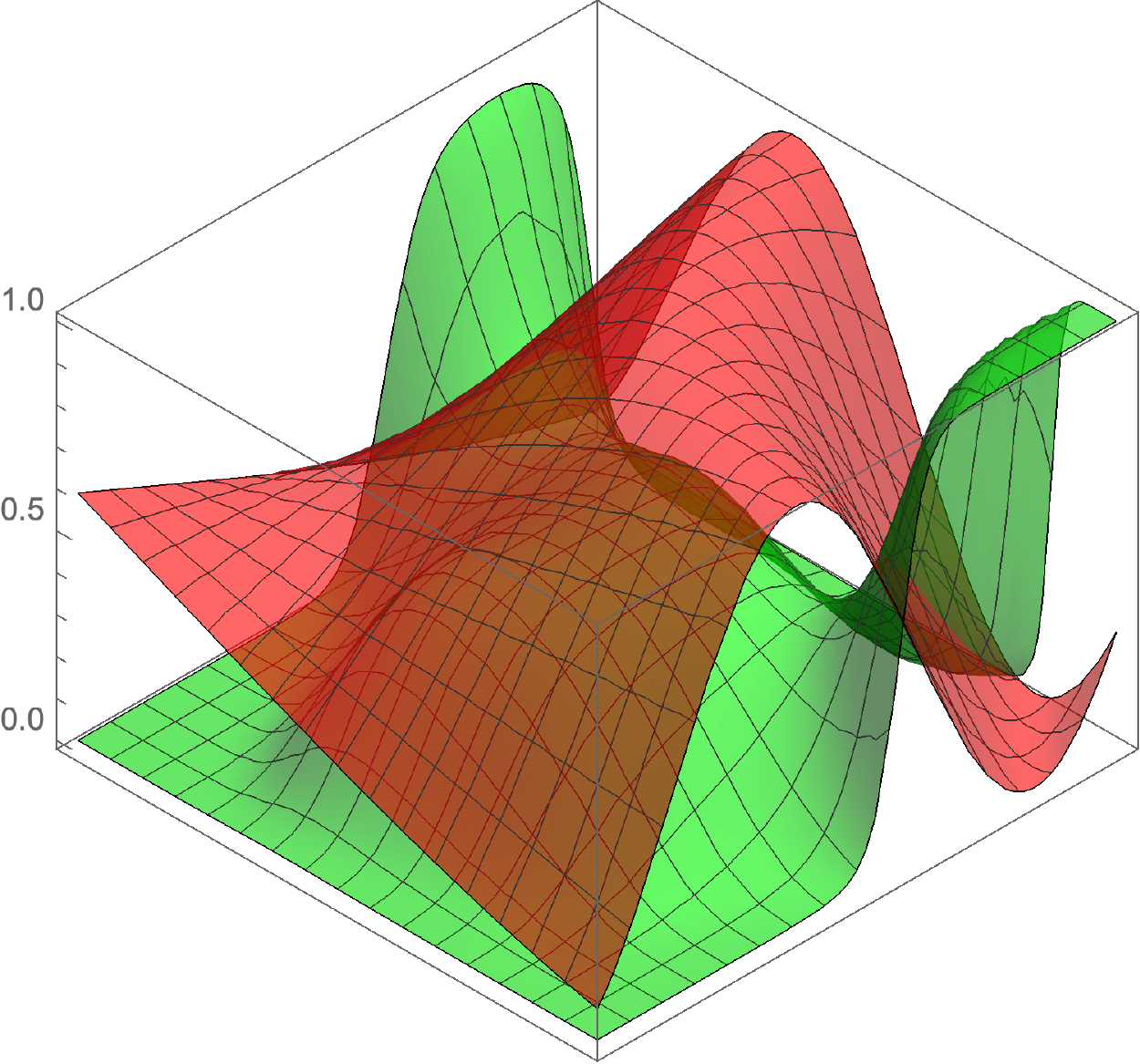}
		\caption{Mixture base}
		\label{fig:mix}
	\end{subfigure}
	\caption{Examples of different $ b $ in two dimensions where green 
	is $ p_{\text{sim}}(E=1 | C=1, \mathbf{x}) $ and red is $ p_{\text{sim}}
	(E=1 | C=0, \mathbf{x})  $}
\end{figure*}

\subsection{Requirements of a simulation}
A number of elements are required in order to simulate a real environment. 
These elements should be taken into account by the signature of a simulation 
and are defined as follows.

\begin{description}
	\item [Drift] In order to test algorithms against drift, in function of time,
	drift must be made a parameter. This can be modeled numerically as $ d(t): t 
	\rightarrow \mathbb{R}^+ $, exerting an influence on a simulation while remaining 
	unknown to the algorithms. If $ d(t) $ fluctuates heavily, a more volatile 
	simulation is created. Less so when $ d(t) $ remains stagnant.
	
	\item [Base functions] Given a single cause situation ($ C \in \{0, 1\} $), 
	while considering a binary effect ($ E \in \{0, 1\} $), four different 
	combinations of $ C $ and $ E $ can be formulated. As such, a simulation 
	should account for these combinations through the aforementioned two 
	distributions $ \mathcal{D}_i $. In their most extreme case, the probabilities 
	associated with $ p(E=1 | C, \mathbf{x}) $ are $ 0 $ and $ 1 $ and must be paired with both 
	states of $ C $. Any interpolation between these probabilities is left to 
	the functional form of the distribution. The complexity of this interpolation 
	will impose a degree of difficulty for the tested algorithm and its method 
	of approximation and should thus be parametrised through what we define as 
	a \textit{base function} $ b $,
	\begin{equation} \label{eq:sim_sign}
		b(C, \mathbf{x}, d(t)): C \times \mathbf{x} \times d(t) \rightarrow [0, 1],
	\end{equation}
	$ \mathbb{D} $ can now be simulated by $ b $ where $ \mathcal{D}_i $ is simulated 
	by $ b(C=i, \mathbf{x}, d(t)) $ which we shall denote $ b_i $.
	
	\item [Effect] An algorithm capable of handling the environment directly, 
	must operate while only receiving some $ \mathbf{x} $ and $ E $ after 
	the application of their chosen $ C $. In this binary effect situation, 
	the simulation should return either $ 1 $ or $ 0 $ representing $ E $, 
	as in reality. As such, any value from $ b $ will be used as a parameter 
	in a Bernoulli experiment.

	\item [Evaluation] The target of an optimal uplift model is to only 
	apply $ C $, when $ \mathbf{x} $ presents a positive causal relationship 
	to $ E $. The notion of this relationship can be presented to the evaluation 
	method as the ground truth is now known through $ b_i $. Furthermore, different 
	intrecacies of the simulation might form additional interest as the evaluation 
	method is highly dependent on the target an algorithm optimises for. With 
	$ b_i $, such intrecacies can be shared to other evaluation methods.
\end{description}

\subsection{Composing the simulation}
We now turn to the problem of composing different $ b_i $. Depending on the 
dimensions describing $ \mathbf{x} \in \mathbb{R}^N $, the domain $ \mathbf{d} 
\in \mathbb{R}^M $ of $ b $ holds different implications. 

When $ N > M $, some dimensions of $ \mathbf{x} $ will have no influence on $ 
\mathcal{D} $, rendering them obsolete. With $ N = M $, every dimension will 
influence $ \mathcal{D} $. In the case of $ N < M $, a group of size $ M - N $ 
unobserved confounders are simulated. To account for an $ M $-dimensional 
domain, a base function $ b $ must be called with some $ \mathbf{x'} = 
(\mathbf{x}, \mathbf{u})^T $, where $ \mathbf{u} \in \mathbb{R}^{M-N} $ 
represents the unobserved confounders. The larger the influence of $ \mathbf{u} $ 
on $ b $ and thus $ \mathcal{D} $, the stronger the influence of the confounder. 
As the name indicates, $ \mathbf{u} $ ought to remain unobservable, i.e. remain 
unknown, to the tested algorithm.

The manner in which the dimensions of $ \mathbf{x'} $ interactively influence 
the simulation is left to $ b $. As such, the method by which $ \mathbf{u} $ 
confounds $ \mathcal{D} $ is another parameter of difficulty for the algorithm, 
though one which we will not further explore in this first formalisation.

While many implementations can be defined for $ b $, we present three different 
approaches, each with its own advantages and disadvantages.
\begin{description}
	\item [Sine base] One approach is to make use of the trigonometric functions 
	(Figure \ref{fig:sine}). This due to their periodic nature, useful for $ d(t) $. 
	Their output range $ [-1, 1] $ is linearly adjusted in (\ref{eq:sine}), with 
	$ \mathbb{I}(\cdot) $ as the indicator function and $ b(C, \mathbf{x'}, d(t)) $ 
	denoted $ b $ for brevity,

	\begin{small} 
		\begin{equation} \label{eq:sine}
			\fontsize{8.5pt}{10.0pt} b_i = \frac{1}{2} \left[ \sin\left( d(t) +
			\lambda_i \prod^M \left[\mathbf{x'}_m + \mathbb{I}(i=0) \mathbf{g}_m 
			\right]\right) +1 \right],	
		\end{equation}		
	\end{small}	
	while drift $ d(t) $ is accounted for through simple summation, interaction 
	in the dimensions of $\mathbf{x'} $ is achieved by multiplication. The 
	complexity of $ b_i $ is parametrised through the addition of a positive integer 
	$ \lambda_i $, governing the frequency of its sine wave such that every cause ($ C=i $) 
	could have a different complexity. A displacement vector $ \mathbf{g} \in \mathbb{R}^M $, 
	monitors the strength of the causal relationship as $ \mathbf{g} $ will provide 
	a difference between $ \mathcal{D}_i $. If $ \mathbf{g} = \mathbf{0} $ no difference 
	is accounted for. A disadvantage of such sine base is the relative ease in which 
	their shapes are estimated.

	\item [Polynomial base] An alternative to the sine base is the polynomial 
	base in (\ref{eq:poly}). Where $ \sigma(\cdot) $ is the logistic sigmoid 
	function, binding the polynomial to a range of $ [0,1] $. The polynomial
	base offers more erratic shapes than the sine base (Figure \ref{fig:poly}), 
	allowing for more rigorous testing. As such, testing using this polynomial 
	base will be more involved with regards to the approximation method.
	\begin{equation} \label{eq:poly}
		b_i = \sigma\left(\left(\mathbf{k}_i + h(d(t))\mathbf{1} \right)^T 
		\bm{\upsilon}_i(\mathbf{x'}) \right)
	\end{equation} 
	With coefficients $ \mathbf{k}_i \in \mathbb{R}^q $, where $ \mathbf{k}_0 
	\neq \mathbf{k}_1 $ and $ q \in \mathbb{N}^+ $, $ h(d(t)) \rightarrow [a, b] $ 
	ensuring some periodic evolution in the interval $ [a, b] $, $ \mathbf{1} $ as 
	a $ q $-dimensional vector with $ \mathbf{1}_j = 1 \text{ } \forall \text{ } 
	j=1,2,...,q $ and
	\begin{equation*}
		\bm{\upsilon}_i(\mathbf{x'}) \doteq \left(1, \prod^1 \upsilon(
		\mathbf{x'}), \prod^2 \upsilon(\mathbf{x'}), ..., \prod^{q-1} 
		\upsilon(\mathbf{x'})\right)^T,
	\end{equation*}
	where $ \bm{\upsilon}_0(\mathbf{x'}) \neq \bm{\upsilon}_1(\mathbf{x'}) $ 
	and $ \upsilon(\cdot) $ will uniformly select one dimension of $ \mathbf{x'} $
	providing a polynomial interaction with a maximum degree of $ q-1 $,
	leaving $ q $ to be a parameter of complexity.

	\item [Prior model base] An argument against above base functions is one of 
	reality. These base functions offer a wide range of complexity, both in
	approximation and in causality, yet their complexity comes entirely from a 
	numerical perspective. A more reality based $ b $ could be a previously 
	trained model, which despite lacking a method to incorporate $ d(t) $ as in 
	(\ref{eq:sim_sign}), could prove useful to relieve a new model from its initial 
	random policy before deployment.
\end{description}

Recall that a causal simulation ought to provide multiple distributions 
$ \mathcal{D}_i $. One strategy to remain independent of a chosen base function 
is to compose a mixture of different base functions for every $ \mathcal{D}_i $ 
(Figure \ref{fig:mix}). Such mixture will provide a wider range of simulations,
yielding a more valuable evaluation.

\subsection{Accounting for robustness}
Probability is a measure of uncertainty governing the occurrence of an event. 
Such uncertainty can be caused by many parameters, though the most reasonable 
one is the unobserved influence of other dimensions. While the explicit introduction 
of unobserved confounders is described above, this section elaborates on 
the implicit introduction of such unobserved confounders through the addition 
of noise.

As such, noise will become another parameter of a simulation. Recall that a 
requirement of a simulation was to provide a strictly binary effect $ E $. 
Therefore, a noisy addition must happen before any Bernoulli experiment takes 
place. While noise can be governed through any distribution, one approach would 
be Gaussian as in (\ref{eq:noise}).
\begin{equation} \label{eq:noise}
	p_{\text{sim}}(E=1 | C, \mathbf{x'}) = \sigma(3P)
\end{equation}

Where $ P \sim \mathcal{N}\left(b(C, \mathbf{x'}, d(t)), \beta^{-1}  \right) $,
$ \beta $ is the precision and $ \mathbb{E}[P] = b(C, \mathbf{x'}, d(t)) $. By 
multiplying $ P $ with a scalar $ 3 $, the logistic sigmoid function can cover 
much wider range of the Gaussian noise. This multiplication is not mandatory and 
can thus be left out. One requirement of the noise is that the expectation should 
equal $ b(C, \mathbf{x'}, d(t)) $ so as to keep its dependency.

\section{Conclusion and further work}
This short paper describes the necessity to test causally aware, experience 
based learners in simulated environments. Arguments for this necessity were 
founded by the shortcomings of data to test concept drift $ d(t) $, the 
dangers of testing experimental algorithms in a real environment and the 
ability to correctly evaluate these algorithms for their causal decisions. 
As such, we introduced a general method of composing such simulation that 
properly tests the robustness of an agent.

Many extensions of this preliminary work can be identified. A first is one of 
state transitions to accommodate a wider range of algorithms. A second extension is 
a thorough analysis of the interaction of $ \mathbf{u} $ to more precisely compose 
both specific situations as evaluations. Following this extension is the wide range 
of possible base functions and their properties that require further investigation.

\section{Appendix}
Documented code composing simulations of above signature can be found 
at \url{https://github.com/vub-dl/cs-um}.
\fontsize{9.3pt}{10.3pt}\selectfont 

\bibliographystyle{aaai}
\end{document}